\PassOptionsToPackage{table}{xcolor}
\documentclass[10pt, a4paper]{article}

\usepackage[table]{xcolor}
\usepackage{subfig}

\usepackage[]{lrec-coling2024} 

\usepackage{graphicx}
\usepackage{tabularx}
\usepackage{soul}

\usepackage{arydshln}

\usepackage{siunitx}
\usepackage{booktabs}
\usepackage{tikz}
\usetikzlibrary{tikzmark, calc}

\usepackage{graphicx}
\usepackage{subfig}
\usepackage{amssymb}
\usepackage{array}
\usepackage{inconsolata}
\usepackage{multirow}
\usepackage[utf8]{inputenc}

\tikzset{
    double color fill/.code 2 args={
        \pgfdeclareverticalshading[%
            tikz@axis@top,tikz@axis@middle,tikz@axis@bottom%
        ]{diagonalfill}{100bp}{%
            color(0bp)=(tikz@axis@bottom);
            color(50bp)=(tikz@axis@bottom);
            color(50bp)=(tikz@axis@middle);
            color(50bp)=(tikz@axis@top);
            color(100bp)=(tikz@axis@top)
        }
        \tikzset{shade, left color=#1, right color=#2, shading=diagonalfill}
    }
}

\newcounter{BGnum}
\usepackage{tikz}
\tikzset{every picture/.style={line width=0pt}}

\usepackage{multirow}

\usepackage{xcolor}
\usepackage{hyperref}
 \definecolor{darkblue}{rgb}{0, 0, 0.5}
  \hypersetup{colorlinks=true, citecolor=darkblue, linkcolor=darkblue, urlcolor=darkblue}

\usepackage{xstring}

\usepackage{color}

\usepackage[11pt]{moresize}

\usepackage{numprint}
\npdecimalsign{.}
\newcommand{\rounded}[1]{\nprounddigits{1}\numprint{#1}}

\newcommand{\roundedbis}[1]{\nprounddigits{2}\numprint{#1}}

\usepackage{float}

\title{How Important Is Tokenization in French Medical Masked Language Models?}

\name{Yanis Labrak{\normalfont \textsuperscript{1,2}}, Adrien Bazoge{\normalfont \textsuperscript{3}} \\ {\large \textbf{ Béatrice Daille{\normalfont \textsuperscript{3}}, Mickael Rouvier{\normalfont \textsuperscript{1}}, Richard Dufour{\normalfont \textsuperscript{1,3}}}} }

\address{{\normalfont \textsuperscript{1}}~LIA, Avignon University \hspace{3mm} {\normalfont \textsuperscript{2}}~Zenidoc \\ {\normalfont \textsuperscript{3}}~Nantes Université, École Centrale Nantes, CNRS, LS2N, UMR 6004, F-44000 Nantes, France \\
         \{first.last\}@\{univ-avignon.fr, univ-nantes.fr\}\\}

\abstract{
Subword tokenization has become the prevailing standard in the field of natural language processing (NLP) over recent years, primarily due to the widespread utilization of pre-trained language models. This shift began with Byte-Pair Encoding (BPE) and was later followed by the adoption of SentencePiece and WordPiece. While subword tokenization consistently outperforms character and word-level tokenization, the precise factors contributing to its success remain unclear. Key aspects such as the optimal segmentation granularity for diverse tasks and languages, the influence of data sources on tokenizers, and the role of morphological information in Indo-European languages remain insufficiently explored. This is particularly pertinent for biomedical terminology, characterized by specific rules governing morpheme combinations. Despite the agglutinative nature of biomedical terminology, existing language models do not explicitly incorporate this knowledge, leading to inconsistent tokenization strategies for common terms. In this paper, we seek to delve into the complexities of subword tokenization in French biomedical domain across a variety of NLP tasks and pinpoint areas where further enhancements can be made. We analyze classical tokenization algorithms, including BPE and SentencePiece, and introduce an original tokenization strategy that integrates morpheme-enriched word segmentation into existing tokenization methods.
\\ \newline  }

\begin{document}

\maketitleabstract

\section{Introduction}

Word tokenization into subword units is a longstanding challenge in the field of natural language processing (NLP), initially conceived to address out-of-vocabulary words in language modeling~\cite{larson2001sub, bazzi2002multi, szoke2008sub}. In recent years, this strategy of splitting words into smaller units has gained prominence, primarily driven by the widespread adoption of pre-trained language models (PLMs) such as BERT~\cite{devlin-etal-2019-bert} and GPT~\cite{NEURIPS2020_1457c0d6}. This shift began with statistical tokenizers, particularly Byte-Pair Encoding (BPE)~\cite{10.5555/177910.177914}, as introduced in BERT~\cite{sennrich-etal-2016-neural}. It was later extended by other data-driven variants, such as SentencePiece (SP)~\cite{kudo-richardson-2018-sentencepiece} and WordPiece~\cite{devlin-etal-2019-bert}.


While empirical evidence consistently demonstrates that subword tokenization outperforms character and word-level tokenization~\cite{10.1007/978-981-10-7134-8_4, wu2016googles}, the precise reasons behind this success remain not fully understood. Some studies have explored the impact of segmentation granularity on subword performance~\cite{samuel-ovrelid-2023-tokenization, novotny-etal-2021-one}, suggesting that each task or language may have its own optimal granularity for maximizing performance. However, other factors, such as the influence of data sources used to construct tokenizers or the role of morphological information, require more comprehensive investigation. 

In the context of Indo-European languages, particularly in French, words are composed of a series of morphemes\footnote{In linguistics, a morpheme is defined as the smallest unit of meaning within a word.}~\cite{touratier_chapitre_2012}. These morphemes can be categorized as either lexical or grammatical and are analogous to the subword units idea previously mentioned, but here adhere to well-defined linguistic rules.

In specialized domains, such as medical one, meaningful morphemes follow construction rules from Greek and Latin languages. These rules help medical professionals deduce the meanings of unfamiliar terms and remember complex terminology effectively. Despite the agglutinative nature of biomedical terminology, existing PLMs do not explicitly integrate this knowledge into their tokenization processes since they only rely on statistical tokenizers (BPE, SentencePiece, etc.). As a result, common terms are inconsistently tokenized into arbitrary subwords in these models.

In this paper, we investigate the impact of word tokenization strategies in the French biomedical domain and their effectiveness on downstream NLP tasks. Our study delves into the nuances of different tokenization algorithms, aiming to understand why subword tokenization strategies, such as BPE and SentencePiece, outperform other methods~\cite{kudo-2018-subword}. We also identify areas for further optimization and provide a comprehensive analysis of their performance on a large set of 23 diverse French biomedical NLP tasks, such as named entity recognition (NER), multi-label classification (CLS), or semantic textual similarity (STS). Finally, we propose an original tokenization strategy that integrates morpheme-enriched word segmentation into existing tokenization algorithms. The latter is included in the comparison of tokenizers and makes it possible to study the contribution of subword units constructed from linguistic rules.

Our contributions are as follows:

\begin{itemize}

\item We introduce an original tokenization strategy that integrates manually defined morphemes into statistical tokenization algorithms.

\item We analyze the ability of statistical tokenizers to segment words regarding their real linguistic segmentation.

\item We provide both qualitative and quantitative analyses to assess how word tokenization approaches (statistical methods vs. morpheme-enriched variants) and the data source on which they are trained impact the performance of BERT-based language models.

\item We explore the relationship between tokenization granularity and its impact on performance in various downstream NLP tasks.


\end{itemize}


The morpheme-enriched tokenization strategy, experiment reproduction scripts, and resulting BERT-based PLMs are freely available under the MIT license on GitHub\footnote{\href{https://github.com/BioMedTok/BioMedTok}{https://github.com/BioMedTok/BioMedTok}} and Hugging Face\footnote{\href{https://huggingface.co/BioMedTok}{https://huggingface.co/BioMedTok}}









\section{Related works}

%
%
%
%

Recent research into domain-specific language models has shown that utilizing specialized data during pre-training significantly enhances model performance in that domain. Various strategies, with varying proportions of in-domain and out-of-domain training data, have been proposed across diverse fields, including biomedicine~\cite{10.1093/bioinformatics/btz682, 10.1145/3458754, el-boukkouri-etal-2022-train, labrak-etal-2023-drbert}, scientific research~\cite{beltagy-etal-2019-scibert} and clinical~\cite{alsentzer-etal-2019-publicly}.

In the context of biomedical-specific language models, it is widely recognized that training models from scratch using in-domain corpora~\cite{10.1145/3458754} yields noticeable performance improvements compared to other pre-training strategies. The authors also demonstrated the benefits of using domain-specific tokenizers generated through conventional statistical tokenization construction techniques, such as WordPiece~\cite{6289079}, SentencePiece~\cite{kudo-richardson-2018-sentencepiece}, and BPE~\cite{sennrich-etal-2016-neural}, on an in-domain corpus, resulting in improved performance in downstream tasks.

Although statistical-based tokenization algorithms are the predominant method employed in recent biomedical language models, some studies have raised questions about the effectiveness of this approach and its suitability for specific downstream tasks or languages~\cite{mielke2021words, novotny-etal-2021-one}. As a result of these inquiries, various methods for improving tokenization have emerged, some involving training models from scratch~\cite{kudo-2018-subword}, while others do not~\cite{hofmann-etal-2022-embarrassingly, fan-sun-2023-constructivist}. One such method involves incorporating linguistic knowledge during the tokenization process by utilizing morphemes~\cite{fujii-etal-2023-different, pan2020morphological, chen-fazio-2021-morphologically, 10.1145/3578707}, with the aim of mimicking how humans learn and understand languages. However, there have been fewer contributions in the context of biomedical domains and Indo-European languages~\cite{jimenez-gutierrez-etal-2023-biomedical}, despite these fields being highly dependent on an agglutinative terminology.

\section{Tokenization Strategies}

In this section, we provide a brief overview of the two studied statistical-based tokenization approaches (Section~\ref{s:tok}), followed by the description of our original approach that integrates linguistic knowledge through morphemes into existing tokenizers algorithms (Section~\ref{s:morph}).

\subsection{Statistical Tokenization Algorithms}
\label{s:tok}


In this study, we compare two statistical-based tokenization methods, BPE and SentencePiece. BPE begins with individual characters and progressively combines them into subword pairs based on their frequency in the training data. In contrast, SentencePiece employs two subword segmentation algorithms, Unigrams and BPE, offering flexibility in terms of segmentation granularity. While SentencePiece is widely used in French biomedical models~\cite{touchent2023camembertbio, labrak-etal-2023-drbert, copara-etal-2020-contextualized, berhe-etal-2023-alibert}, its appropriateness for a specific language and domain may vary, potentially leading to suboptimal subword segmentation.

\subsection{Morpheme-enriched Tokenization}
\label{s:morph}


In our study, focusing on improving the modeling of specialized medical terminology in the medical field and reducing the impact of unseen words during model pre-training, our primary emphasis is on lexical morphemes~\cite{touratier_chapitre_2012}. To achieve this, we created a manual list of around 600 frequently used lexical morphemes in the French medical domain, sourced from the book by \citet{cottez1980dictionnaire}. Examples of these morphemes include terms like \texttt{céphal-}, \texttt{clinico-}, \texttt{-thérapie}, \texttt{thoraco-}, \texttt{-ome} and \texttt{-gène}.




We trained our morpheme-enriched tokenizers by modifying both the BPE and SentencePiece algorithms. During training, we introduced a predefined list of language-specific morphemes as tokens. These morphemes were enforced selections by the tokenizer when encountered, while the remaining text underwent the standard tokenization process of the chosen algorithm. This approach enabled us to combine traditional BPE and SentencePiece tokenizations with morpheme tokens, mitigating issues related to unseen words during training.

\section{Experimental Protocol}

In this section, we outline the experimental approach used to evaluate the impact of tokenization strategies on French biomedical PLMs. Firstly, in Section~\ref{sec:downstream}, we present the set of 23 selected biomedical NLP downstream tasks used in our study. Next, we describe the different training data sources employed to train the statistical tokenizers in Section~\ref{sec:influence-tok-data}. Following this, in Section~\ref{sec:lm-training}, we explain the training procedure for the chosen BERT-based model architecture. Finally, in Section~\ref{sec:evaluation}, we provide a comprehensive description of the evaluation methodology used to assess the performance of these models.

\subsection{Downstream Tasks}
\label{sec:downstream}

We summarize the datasets of the 23 NLP biomedical downstream tasks from DrBenchmark~\cite{labrak2024drbenchmark}, including NER, part-of-speech (POS) tagging, STS and classification.

\paragraph{DEFT-2020}~\citelanguageresource{cardon-etal-2020-presentation} is a dataset featured in the 2020 edition of the annual French Text Mining Challenge, known as DEFT. It encompasses clinical cases, encyclopedia, and drug labels, all of which have been annotated for two specific tasks: (i) assessing textual similarity and (ii) performing multi-class classification. The first task is geared towards determining the degree of similarity between pairs of sentences, with a scale ranging from 0 to 5 and involves 1,010 sample pairs. The second task involves identifying, for a given sentence, which among three provided sentences is the most similar. There are 1,102 samples included in this task.

\paragraph{DEFT-2021}~\citelanguageresource{grouin-etal-2021-classification} is a subset of 275 clinical cases taken from the 2019 edition of DEFT. This dataset is manually annotated in two tasks: (i) multi-label classification with 275 samples and (ii) NER. The multi-label classification task focuses on identifying the patient's clinical profile based on the diseases, signs, or symptoms mentioned in the clinical cases with 4,712 samples. The dataset is annotated with 23 axes derived from Chapter C of the Medical Subject Headings (MeSH). The second task involves fine-grained information extraction for 13 entities.





\paragraph{E3C}~\citelanguageresource{Magnini2020TheEP} is a multilingual collection of clinical cases annotated for Named Entity Recognition (NER). It encompasses two types of annotations: (i) clinical entities and (ii) temporal information and factuality. While this dataset spans five languages, our evaluation focuses on the French portion. Since the dataset does not come with predefined subsets for its 1,402 samples, we conducted random splits of 70\% for training, 10\% for validation, and 20\% for testing, as outlined in Table~\ref{table:sources-e3c}.

\begin{table}[H]
\centering
\resizebox{\columnwidth}{!}{%
\begin{tabular}{|l|lll|}
\hline
\textbf{Subset}            & \textbf{Train} & \textbf{Validation} & \textbf{Test} \\ \hline
\textit{\textbf{Clinical}} & 87.38\% of L2  & 12.62\% of L2       & 100\% of L1   \\
\textit{\textbf{Temporal}} & 70\% of L1     & 10\% of L1          & 20\% of L1    \\ \hline
\end{tabular}%
}
\caption{Description of the sources for E3C.}
\label{table:sources-e3c}
\end{table}

\paragraph{The QUAERO French Medical Corpus}~\citelanguageresource{Nvol2014TheQF}, simply referred to as QUAERO in this paper, contains annotated entities and concepts for NER tasks. The dataset covers two text genres (drug leaflets and biomedical titles). 10 entity categories corresponding to the UMLS Semantic Groups~\citelanguageresource{Lindberg-MIM1993} were annotated, for a total of 26,409 entity, which were mapped to 5,797 unique UMLS concepts. 
Due to the presence of nested entities, we opted to simplify the evaluation process by retaining only annotations at the higher granularity level, following a similar approach to the one described in~\citet{touchent2023camembertbio}, which translates into an average loss of 6.06\% of the annotations on EMEA and 8.90\% on MEDLINE. Additionally, considering that some documents from EMEA exceed the maximum input sequence length that most current language models can handle, we decided to split these documents into sentences.

\paragraph{MorFITT}~\citelanguageresource{labrak:hal-04125879} is a multi-label dataset that has been annotated with medical specialties. It comprises 3,624 biomedical abstracts sourced from PMC Open Access. These abstracts have been annotated across 12 distinct medical specialties, resulting in a total of 5,116 annotations.


\paragraph{Mantra-GSC}~\citelanguageresource{10.1093/jamia/ocv037} is a multilingual dataset annotated in biomedical NER for five languages, however we focused only on the French subset. It covers three sources (EMEA, Medline and Patents) and use two distinct annotation schemes. These sources encompass diverse types of documents, including biomedical abstracts/titles, drug labels, and patents. To maintain evaluation uniformity, we randomly divided the dataset into three subsets: 70\% for training, 10\% for validation, and 20\% for testing.


\paragraph{CLISTER}~\citelanguageresource{hiebel-etal-2022-clister-corpus} is a collection of French clinical case sentence pairs used for Semantic Textual Similarity (STS) evaluation. It consists of 1,000 sentence pairs, manually annotated by multiple annotators who assigned similarity scores ranging from 0 to 5 for each pair. These individual scores were then averaged to derive a floating-point number that represents the overall similarity of the two sentences.



\paragraph{CAS}~\citelanguageresource{grabar:hal-01937096} dataset comprises 3,790 clinical cases that underwent POS tagging with 31 different classes, using automatic tagging through the Tagex tool\footnote{\href{https://allgo.inria.fr/app/tagex}{https://allgo.inria.fr/app/tagex}}, achieving a 98\% precision rate in comparison to manual annotations. This dataset involves tasks like classifying clinical cases for negation and uncertainty, as well as named-entity recognition for identifying markers of negations and speculation within medical histories and patient care. To create subsets, a random split was applied, allocating 70\% for training, 10\% for validation, and 20\% for testing since predefined subsets were not provided.

\paragraph{ESSAI}~\citelanguageresource{dalloux_claveau_grabar_oliveira_moro_gumiel_carvalho_2021} consists of 7,247 clinical trial protocols that have been annotated with 41 POS tags using the TreeTagger tool~\citelanguageresource{Schmid1994}. It does also contain a classification and two named-entity recognition tasks similar to those from CAS dataset. As the dataset was not initially separated into three distinct subsets, we opted to apply the same processing methodology as we did for CAS dataset.

\paragraph{PxCorpus}~\citelanguageresource{Kocabiyikoglu2022} is a dataset designed for spoken language understanding in the medical domain, specifically focusing on transcripts related to drug prescriptions. It comprises 4 hours of transcribed dialogues, amounting to 1,981 recordings. These dialogues have been meticulously transcribed and semantically annotated. The primary task involves categorizing the textual utterances into one of four intent classes (prescribe, replace, negate, none). The second task pertains to NER, where each word in a sequence is classified into one of 38 classes, including categories such as drug, dose, or mode.

\subsection{Tokenizers Data Sources}
\label{sec:influence-tok-data}

To ensure a fair and comprehensive comparison of training data sources used by the statistical tokenizers, we carefully curated a 1GB subset of raw, lowercase text data from a variety of sources, including NACHOS~\citelanguageresource{labrak-etal-2023-drbert}, PubMed Central, CC100~\citelanguageresource{wenzek-etal-2020-ccnet}, and the French Wikipedia. We then constructed tokenizers using both tokenization algorithms, resulting in a total of 16 tokenizers: 8 with the integration of morphemes and 8 without. These specific data sources were chosen for their diversity: NACHOS focuses on French biomedical content, PubMed Central on English biomedical content, Wikipedia on general French language, and CC100 on general multilingual content. Each tokenizer was configured with a vocabulary size of 32k tokens, consistent with the original hyperparameters used in other French biomedical models such as CamemBERT-BIO~\citelanguageresource{touchent2023camembertbio} and DrBERT~\citelanguageresource{labrak-etal-2023-drbert}.

\subsection{Language Model Pre-Training}
\label{sec:lm-training}


To assess the impact of introducing morphemes into tokenizers on the pre-training process of biomedical language models, we conducted pre-training from scratch using the 16 tokenizer combinations (see Section~\ref{sec:influence-tok-data}). Our choice of architecture was RoBERTa~\cite{liu2020roberta}, which is based on the masked language modeling objective and configured with standard token masking percentages as introduced by the authors.

For the PLMs training data, we utilized the NACHOS corpus created by~\cite{labrak-etal-2023-drbert}. This corpus, already pre-processed and converted to lowercase, is consistent with the data sources used for training the tokenizers. It comprises 1.1 billion words, equivalent to 7.4GB of raw text data, sourced from a wide range of online resources focusing on the French biomedical and clinical domains.

The pre-training process was conducted uniformly across all models, employing the same hyperparameters and executed over a 20-hour period. We harnessed the computational power of 32 V100 32GB GPUs available on the Jean-Zay supercomputer for this purpose. By maintaining consistent procedures and employing a fixed seed to mitigate randomness during training, we ensured the reliability and reproducibility of our experiments.

\begin{table*}[!t]
\centering
\setlength\tabcolsep{1.7pt}
\setlength\extrarowheight{2.5pt}
\fontsize{6.5pt}{6.5pt}\selectfont
\begin{tabular}{ccc|cccccccc|cccccccc|}

\cline{4-19}

 &
   &
   &
  \multicolumn{8}{c|}{\textbf{BPE}} &
  \multicolumn{8}{c|}{\textbf{SentencePiece}} \\ \cline{4-19} 
  
 &
   &
   &
  \multicolumn{2}{c}{\textbf{\textit{NACHOS}}} &
  \multicolumn{2}{c}{\textbf{\textit{PubMed}}} &
  \multicolumn{2}{c}{\textbf{\textit{CC100}}} &
  \multicolumn{2}{c|}{\textbf{\textit{Wiki}}} &
  \multicolumn{2}{c}{\textbf{\textit{NACHOS}}} &
  \multicolumn{2}{c}{\textbf{\textit{PubMed}}} &
  \multicolumn{2}{c}{\textbf{\textit{CC100}}} &
  \multicolumn{2}{c|}{\textbf{\textit{Wiki}}} \\ \hline
  
\multicolumn{1}{|c}{\textbf{Dataset}} &
  \textbf{Task} &
  \textbf{Metric} &
  \textit{w/o} &
  \textit{w/} &
  
  \textit{w/o} &
  \textit{w/} &
  
  \textit{w/o} &
  \textit{w/} &
  
  \textit{w/o} &
  \textit{w/} &
  
  \textit{w/o} &
  \textit{w/} &
  
  \textit{w/o} &
  \textit{w/} &
  
  \textit{w/o} &
  \textit{w/} &
  
  \textit{w/o} &
  \textit{w/}
  
  \\ \hline
\multicolumn{1}{|c}{\multirow{4}{*}{\textbf{CAS}}} &
  \textbf{\textit{CLS}} &
  F1 &
  \rounded{94.21}* &
  \rounded{94.92} &
  \rounded{94.66} &
  \rounded{94.18}* &
  \rounded{95.20} &
  \textbf{\rounded{95.32}} &
  \rounded{94.83} &
  \rounded{94.75} &
  \rounded{94.79} &
  \rounded{94.69}* &
  \rounded{93.37}** &
  \rounded{93.57}** &
  \rounded{94.42} &
  \rounded{94.11} &
  \underline{\rounded{95.25}}* &
  \rounded{95.05}** \\
\multicolumn{1}{|c}{} &
  \textbf{\textit{NER Neg}} &
  SeqEval &
  \textbf{\rounded{87.01}} &
  \rounded{83.30}* &
  \rounded{82.38}** &
  \rounded{81.32}** &
  \rounded{84.85} &
  \rounded{84.23} &
  \rounded{84.70} &
  \rounded{84.51}* &
  \rounded{86.07} &
  \underline{\rounded{86.42}} &
  \rounded{83.61}* &
  \rounded{83.87} &
  \rounded{85.40} &
  \rounded{84.16}** &
  \rounded{85.62} &
  \rounded{83.19} \\
  
\multicolumn{1}{|c}{} &
  \textbf{\textit{NER Spec}} &
  SeqEval  &
  \rounded{30.30}* &
  \rounded{30.56} &
  \underline{\rounded{35.04}} &
  \rounded{28.21}* &
  \rounded{34.58} &
  \rounded{32.00} &
  \rounded{34.38} &
  \rounded{34.02} &
  \textbf{\rounded{36.05}} &
  \rounded{29.82} &
  \rounded{28.42}* &
  \rounded{22.18}** &
  \rounded{31.94} &
  \rounded{28.72}* &
  \rounded{32.07} &
  \rounded{26.99}* \\
  
\multicolumn{1}{|c}{} &
  \textbf{\textit{POS}} &
  SeqEval  &
  \rounded{96.98}** &
  \rounded{96.88}** &
  \rounded{97.07} &
  \rounded{96.86}** &
  \rounded{97.06}* &
  \rounded{96.96}** &
  \textbf{\rounded{97.16}} &
  \rounded{96.94}** &
  \rounded{97.05}* &
  \rounded{97.02}* &
  \rounded{96.94}** &
  \rounded{96.87}** &
  \rounded{97.12} &
  \underline{\rounded{97.14}} &
  \rounded{97.12} &
  \rounded{97.12} \\ \hline
\multicolumn{1}{|c}{\multirow{2}{*}{\textbf{PxCorpus}}} &
  \textbf{\textit{CLS}} &
    F1 &
  \underline{\rounded{94.82}} &
  \rounded{94.22} &
  \rounded{93.64} &
  \rounded{93.91} &
  \rounded{94.24} &
  \rounded{94.58} &
  \rounded{93.39} &
  \rounded{93.65} &
  \textbf{\rounded{94.92}} &
  \rounded{94.07} &
  \rounded{94.78} &
  \rounded{94.09} &
  \rounded{94.76} &
  \rounded{93.66} &
  \rounded{93.73} &
  \rounded{94.48} \\
\multicolumn{1}{|c}{} &
  \textbf{\textit{NER}} &
  SeqEval  &
  \rounded{95.86} &
  \rounded{95.85} &
  \rounded{95.93} &
  \rounded{95.87} &
  \rounded{96.14} &
  \rounded{95.99} &
  \underline{\rounded{96.17}} &
  \rounded{95.94} &
  \rounded{96.14} &
  \rounded{96.12} &
  \rounded{95.96} &
  \rounded{96.07} &
  \rounded{95.93} &
  \rounded{96.06} &
  \textbf{\rounded{96.23}} &
  \rounded{96.11} \\ \hline
\multicolumn{1}{|c}{\multirow{2}{*}{\textbf{DEFT2020}}} &
  \textbf{\textit{STS}} &
  MSE &
  0.71 &
  0.71 &
  0.64* &
  \textbf{0.75} &
  0.70 &
  0.67 &
  0.71 &
  0.69 &
  \underline{0.72} &
  0.71 &
  0.63** &
  0.63 &
  0.70 &
  0.67* &
  0.70 &
  0.67* \\
\multicolumn{1}{|c}{} &
  \textbf{\textit{CLS}} &
  F1 &
  \textbf{\rounded{90.98}} &
  \underline{\rounded{85.93}} &
  \rounded{57.60}** &
  \rounded{73.69} &
  \rounded{79.52} &
  \rounded{76.28} &
  \rounded{77.12} &
  \rounded{65.95} &
  \rounded{82.95} &
  \rounded{85.29} &
  \rounded{80.87} &
  \rounded{66.70}** &
  \rounded{61.08}* &
  \rounded{66.30}* &
  \rounded{74.97}* &
  \rounded{77.39}* \\ \hline
\multicolumn{1}{|c}{\textbf{MORFITT}} &
  \textbf{\textit{CLS}} &
  F1 &
  \rounded{68.55}** &
  \rounded{67.97}** &
  \rounded{66.50}** &
  \rounded{65.92}** &
  \rounded{68.43}** &
  \rounded{67.04}** &
  \rounded{68.74} &
  \rounded{67.31}** &
  \textbf{\rounded{69.64}} &
  \rounded{68.77}* &
  \rounded{66.76}** &
  \rounded{66.19}** &
  \rounded{68.22} &
  \rounded{67.48}** &
  \underline{\rounded{69.07}}** &
  \rounded{67.71}** \\ \hline
\multicolumn{1}{|c}{\multirow{2}{*}{\textbf{E3C}}} &
  \textbf{\textit{NER Clinical}} &
  SeqEval &
  \textbf{\rounded{54.18}} &
  \rounded{53.05} &
  \rounded{52.35} &
  \rounded{48.60}** &
  \rounded{52.71} &
  \rounded{51.33}** &
  \rounded{51.13}* &
  \rounded{52.02}* &
  \underline{\rounded{54.17}} &
  \rounded{52.37} &
  \rounded{52.06} &
  \rounded{51.08}** &
  \rounded{53.79} &
  \rounded{52.46}* &
  \rounded{53.21} &
  \rounded{51.73} \\
\multicolumn{1}{|c}{} &
  \textbf{\textit{NER Temporal}} &
  SeqEval &
  \rounded{82.01} &
  \rounded{81.16} &
  \rounded{80.91}** &
  \rounded{80.03}** &
  \rounded{81.82} &
  \rounded{81.22} &
  \textbf{\rounded{82.32}} &
  \rounded{80.62}** &
  \underline{\rounded{82.10}} &
  \rounded{81.60} &
  \rounded{80.30}** &
  \rounded{79.80}** &
  \rounded{80.63}** &
  \rounded{81.11}** &
  \rounded{81.62}* &
  81.73* \\ \hline
\multicolumn{1}{|c}{\textbf{CLISTER}} &
  \textbf{\textit{STS}} &
  MSE &
  0.63* &
  0.63 &
  0.63 &
  0.60** &
  \underline{0.65} &
  0.63 &
  0.62** &
  \textbf{0.66} &
  0.61* &
  0.64 &
  0.61** &
  0.62** &
  0.62 &
  0.60* &
  0.64** &
  0.63** \\ \hline
\multicolumn{1}{|c}{\multirow{2}{*}{\textbf{DEFT2021}}} &
  \textbf{\textit{NER}} &
  SeqEval &
  \underline{\rounded{60.32}} &
  \rounded{58.96}** &
  \rounded{58.06}** &
  \rounded{56.21}** &
  \rounded{59.39}** &
  \rounded{59.20}** &
  \rounded{60.06}** &
  \rounded{59.05}** &
  \textbf{\rounded{61.25}} &
  \rounded{60.09}* &
  \rounded{57.00}** &
  \rounded{56.58}** &
  \rounded{59.21}** &
  \rounded{59.90}** &
  \rounded{59.31}** &
  \rounded{58.94}** \\
\multicolumn{1}{|c}{} &
  \textbf{\textit{CLS}} &
  F1 &
  \rounded{32.89} &
  \underline{\rounded{34.51}}* &  
  \rounded{33.38} &
  \rounded{32.34} &  
  \rounded{34.46}* &
  \rounded{33.94} &  
  \rounded{34.19} &
  \rounded{32.89} &  
  \rounded{34.27} &
  \rounded{33.07} &  
  \rounded{34.30} &
  \rounded{33.13} &  
  \rounded{30.95} &
  \rounded{31.88}* &  
  \rounded{34.15} &
  \textbf{\rounded{34.86}} \\ \hline
  
\multicolumn{1}{|c}{\multirow{4}{*}{\textbf{ESSAI}}} &
  \textbf{\textit{NER Spec}} &
  SeqEval &
  \rounded{60.47} &
  \rounded{60.93} &  
  \rounded{56.41}* &
  \rounded{59.21} &  
  \rounded{57.93} &
  \rounded{61.46} &  
  \rounded{63.56} &
  \rounded{57.43} &  
  \underline{\rounded{63.88}} &
  \rounded{62.78} &  
  \rounded{57.63} &
  \rounded{55.68}* &  
  \textbf{\rounded{64.58}} &
  \rounded{61.97} &  
  \rounded{61.43} &
  \rounded{63.06} \\
  
\multicolumn{1}{|c}{} &
  \textbf{\textit{POS}} &
  SeqEval &
  \underline{\rounded{98.39}}* &
  \rounded{98.29} &  
  \rounded{98.34} &
  \rounded{98.22}** &  
  \rounded{98.37} &
  \rounded{98.38} &  
  \rounded{98.33} &
  \rounded{98.28} &  
  \rounded{98.35} &
  \rounded{98.35} &  
  \rounded{98.30} &
  \rounded{98.21}* &  
  \textbf{\rounded{98.4}} &
  \rounded{98.32} &
  \rounded{98.32}* &
  \rounded{98.31} \\

\multicolumn{1}{|c}{} &
  \textbf{\textit{NER Neg}} &
  SeqEval &
  \rounded{83.02} &
  \rounded{83.42} &
  \rounded{79.26} &
  \rounded{76.40} &
  \rounded{82.20} &
  \rounded{83.18} &
  \rounded{81.81} &
  \underline{\rounded{84.22}}* &
  \rounded{81.29} &
  \rounded{83.99}* &
  \rounded{80.22} &
  \rounded{81.06} &
  \rounded{83.18} &
  \textbf{\rounded{84.24}} &
  \rounded{82.11} &
  \rounded{79.63}* \\
  
\multicolumn{1}{|c}{} &
  \textbf{\textit{CLS}} &
  F1 &
  \rounded{97.31} &
  \rounded{97.06}* &
  \rounded{97.44} &
  \rounded{96.56}** &
  \rounded{97.37} &
  \rounded{96.71}** &
  \rounded{97.38} &
  \rounded{97.00}** &
  \rounded{97.29} &
  \rounded{97.32} &
  \textbf{\rounded{97.49}} &
  \rounded{97.16}* &
  \rounded{97.04} &
  \rounded{97.04} &
  \underline{\rounded{97.48}}* &
  \rounded{97.04}* \\ \hline
  
\multicolumn{1}{|c}{\multirow{2}{*}{\textbf{QUAERO}}} &
  \textbf{\textit{NER Medline}} &
  SeqEval &
  \rounded{57.66} &
  \rounded{56.18}** &
  \rounded{55.44}** &
  \rounded{53.63}** &
  \underline{\rounded{57.85}} &
  \rounded{55.00}** &
  \rounded{57.32} &
  \rounded{56.36}** &
  \textbf{\rounded{58.18}} &
  \rounded{55.47}** &
  \rounded{54.82}** &
  \rounded{52.89}** &
  \rounded{57.54}* &
  \rounded{55.83}** &
  \rounded{56.91} &
  \rounded{54.88}** \\
  
\multicolumn{1}{|c}{} &
  \textbf{\textit{NER EMEA}} &
  SeqEval &
  \underline{\rounded{65.63}} &
  \rounded{65.12} &
  \rounded{63.88} &
  \rounded{63.10}** &
  \rounded{62.10}** &
  \rounded{62.68}* &
  \rounded{63.05}** &
  \rounded{62.63}* &
  \rounded{65.54} &
  \textbf{\rounded{65.89}} &
  \rounded{62.59}** &
  \rounded{63.79}* &
  \rounded{62.84}** &
  \rounded{63.09}* &
  \rounded{62.74}* &
  \rounded{61.99}** \\ \hline
  
\multicolumn{1}{|c}{\multirow{3}{*}{\textbf{MantraGSC}}} &
  \textbf{\textit{NER EMEA}} &
  SeqEval &
  \rounded{60.94} &
  \rounded{63.92} &
  \rounded{58.15}* &
  \rounded{60.56}* &
  \textbf{\rounded{69.34}} &
  \rounded{62.98} &
  \rounded{61.87}* &
  \rounded{62.29}** &
  \underline{\rounded{66.86}} &
  \rounded{62.54}* &
  \rounded{56.79}** &
  \rounded{60.32} &
  \rounded{60.78}* &
  \rounded{59.53} &
  \rounded{63.95}* &
  \rounded{63.85}** \\
  
\multicolumn{1}{|c}{} &
  \textbf{\textit{NER Medline}} &
  SeqEval &
  \rounded{41.36}* &
  \rounded{42.94} &
  \rounded{39.27} &
  \rounded{36.22}** &
  \rounded{44.30} &
  \rounded{41.16} &
  \rounded{43.76} &
  \rounded{40.79}* &
  \rounded{41.87} &
  \rounded{39.47}* &
  \rounded{36.36}** &
  \rounded{37.84} &
  \underline{\rounded{46.37}}* &
  \rounded{39.86} &
  \textbf{\rounded{47.14}} &
  \rounded{36.10}* \\
  
\multicolumn{1}{|c}{} &
  \textbf{\textit{NER Patents}} &
  SeqEval &
  \rounded{52.07}* &
  \rounded{53.32}* &
  \textbf{\rounded{57.03}} &
  \rounded{50.18}* &
  \underline{\rounded{56.95}} &
  \rounded{53.93} &
  \rounded{53.64} &
  \rounded{52.33}* &
  \rounded{51.98} &
  \rounded{49.62}* &
  \rounded{50.74}** &
  \rounded{49.44} &
  \rounded{52.83}* &
  \rounded{48.04} &
  \rounded{50.55}* &
  \rounded{47.76}* \\
  
  \hline

\multicolumn{19}{|c|}{\textit{Average performances per tasks}} \\
  
  \hline

\multicolumn{2}{|c}{\textbf{CLS}} & F1 &
  \textbf{79.80} & 79.10 & \underline{73.87} & 76.10 & 78.20 & 77.30 & 77.60 & 75.28 & 78.98 & 78.88 & 77.95 & 75.15 & 74.42 & 75.08 & 77.47 & 77.77 \\
\multicolumn{2}{|c}{\textbf{NER}} & SeqEval &
  63.92 & 63.75 & 62.63 & \underline{60.73} & 64.63 & 63.42 & 64.15 & 63.24 & \textbf{65.05} & 63.55 & 61.27 & 60.82 & 64.22 & 62.69 & 64.06 & 62.00 \\
\multicolumn{2}{|c}{\textbf{POS}} & SeqEval &
  97.70 & 97.60 & 97.70 & \underline{97.55} & \textbf{97.75} & 97.70 & 97.75 & 97.60 & 97.75 & 97.70 & 97.60 & 97.55 & 97.75 & 97.70 & 97.70 & 97.70 \\
\multicolumn{2}{|c}{\textbf{STS}} & MSE &
  0,67 & 0,67 & 0,64 & 0,68 & \textbf{0,68} & 0,65 & 0,67 & 0,68 & 0,67 & 0,68 & \underline{0,62} & 0,63 & 0,66 & 0,64 & 0,67 & 0,65 \\ \hline


\end{tabular}%
\caption{Performance of the tokenization algorithms and different data sources used to train tokenizers (top). Average performance per type of tasks is also reported (bottom). {\it w/o} and {\it w/} denote models without and with morphemes. Best models are  in bold, and the second-best are underlined. Statistical significance is determined using Student’s t-test, where * indicates p < 0.05, and **  p < 0.01.}
\label{table:results}
\end{table*}

\subsection{Evaluation}
\label{sec:evaluation}



All models undergo fine-tuning following a standardized protocol with identical hyperparameters for each downstream task, enabling a focused evaluation of tokenizers. We ensure robustness and reliability by averaging the results across four independent runs and performing statistical significance assessments using Student’s t-test.

For consistent comparisons, especially in sequence-to-sequence tasks like POS tagging and NER, we employ the SeqEval~\citelanguageresource{seqeval} metric in conjunction with the IOB2 format. To align with established practices~\cite{touchent2023camembertbio}, our models are trained to predict only the label for the initial token of each word.
\section{Results and Discussions}


In this section, we present the results of our tokenization strategies on various biomedical NLP tasks, with a focus on key aspects. We investigate the impact of tokenization granularity (Section~\ref{sec:impact-token-granularity}), the introduction of morphological information during tokenizer construction (Section~\ref{sec:morpheme-impact}), and the influence of data sources on tokenizers, including token sparsity, morpheme coverage, and the overall performance of different tokenization algorithms (Section~\ref{sec:impact-data-sources}).

Table~\ref{table:results} summarizes the performance of the BPE and SentencePiece strategies, both with ({\it w/}) and without our morpheme-enriched approach ({\it w/o}), across various French biomedical downstream tasks. Average performance per task type is also provided for clarity. It's worth noting that, before delving into detailed analysis, there is no consistent tokenization strategy that consistently yields the best results in all tasks, whether it employs a purely statistical algorithm or a statistical approach coupled with morpheme enrichment.

\begin{table*}[!htb]
\setlength\tabcolsep{2.4pt}
\setlength\extrarowheight{1.0pt}
\scriptsize
\centering
\begin{tabular}{cc|cccccccc|cccccccc||c}

\cline{3-18}

&
&
\multicolumn{8}{c|}{\textbf{BPE}} &
\multicolumn{8}{c|}{\textbf{SentencePiece}} \\ 
\cline{3-18}

&
&
\multicolumn{2}{c|}{\textbf{\textit{NACHOS}}} &
\multicolumn{2}{c|}{\textbf{\textit{PubMed}}} &
\multicolumn{2}{c|}{\textbf{\textit{CC100}}} &
\multicolumn{2}{c|}{\textbf{\textit{Wikipedia}}} &

\multicolumn{2}{c|}{\textbf{\textit{NACHOS}}} &
\multicolumn{2}{c|}{\textbf{\textit{PubMed}}} &
\multicolumn{2}{c|}{\textbf{\textit{CC100}}} &
\multicolumn{2}{c|}{\textbf{\textit{Wikipedia}}} \\ 
\cline{1-19} 

\multicolumn{1}{|c}{\textbf{Corpus}} &
\textbf{Task} &
 \textit{w/o} &\textit{ w/} &
 \textit{w/o} &\textit{ w/} &
 \textit{w/o} &\textit{ w/} &
 \textit{w/o} &\textit{ w/} &
 \textit{w/o} &\textit{ w/} &
 \textit{w/o} &\textit{ w/} &
 \textit{w/o} &\textit{ w/} &
 \textit{w/o} &\textit{ w/} &
 \multicolumn{1}{c|}{\textit{$\rho$}}
 \\ \hline
 

\multicolumn{1}{|c}{\multirow{4}{*}{\textbf{CAS}}}
 & \textbf{CLS} & \roundedbis{1.32} & \roundedbis{1.38} & \roundedbis{2.2} & \roundedbis{2.13} & \roundedbis{1.49} & \cellcolor{green!25} \roundedbis{1.49} & \roundedbis{1.51} & \roundedbis{1.5} & \roundedbis{1.32} & \roundedbis{1.45} & \cellcolor{red!25} \roundedbis{2.18} & \roundedbis{2.15} & \roundedbis{1.49} & \roundedbis{1.56} & \roundedbis{1.51} & \roundedbis{1.57} & \multicolumn{1}{c|}{\textit{-0.62}}  \\
 \multicolumn{1}{|c}{} & \textbf{NER Neg} & \cellcolor{green!25}\roundedbis{1.32} & \roundedbis{1.38} & \roundedbis{2.2} & \cellcolor{red!25} \roundedbis{2.13} & \roundedbis{1.49} & \roundedbis{1.49} & \roundedbis{1.51} & \roundedbis{1.5} &  \roundedbis{1.32} & \roundedbis{1.45} & \roundedbis{2.18} & \roundedbis{2.15} & \roundedbis{1.49} & \roundedbis{1.56} & \roundedbis{1.51} & \roundedbis{1.57} & \multicolumn{1}{c|}{\textit{-0.70}} \\
\multicolumn{1}{|c}{}  & \textbf{NER Spec} & \roundedbis{1.32} & \roundedbis{1.38} & \roundedbis{2.2} & \roundedbis{2.13} & \roundedbis{1.49} & \roundedbis{1.49} & \roundedbis{1.51} & \roundedbis{1.5} & \cellcolor{green!25} \roundedbis{1.32} & \roundedbis{1.45} & \roundedbis{2.18} & \cellcolor{red!25} \roundedbis{2.15} & \roundedbis{1.49} & \roundedbis{1.56} & \roundedbis{1.51} & \roundedbis{1.57} & \multicolumn{1}{c|}{\textit{-0.42}} \\
\multicolumn{1}{|c}{} & \textbf{POS} & \roundedbis{1.32} & \cellcolor{red!25} \roundedbis{1.38} & \roundedbis{2.2} & \cellcolor{red!25} \roundedbis{2.13} & \roundedbis{1.49} & \roundedbis{1.49} & \cellcolor{green!25} \roundedbis{1.51} & \cellcolor{red!25} \roundedbis{1.5} & \roundedbis{1.32} & \roundedbis{1.45} & \cellcolor{red!25} \roundedbis{2.18} &  \cellcolor{red!25}\roundedbis{2.15} & \roundedbis{1.49} & \roundedbis{1.56} & \roundedbis{1.51} & \roundedbis{1.57} & \multicolumn{1}{c|}{\textit{-0.36}} \\ \hline

\multicolumn{1}{|c}{\multirow{2}{*}{\textbf{PxCorpus}}}
 & \textbf{CLS} & \cellcolor{green!25} \roundedbis{1.54} & \roundedbis{1.62} & \cellcolor{red!25} \roundedbis{2.26} & \roundedbis{2.27} & \roundedbis{1.76} & \roundedbis{1.72} & \roundedbis{1.73} & \roundedbis{1.72} & \roundedbis{1.54} & \roundedbis{1.67} & \roundedbis{2.24} & \roundedbis{2.3} & \roundedbis{1.72} & \roundedbis{1.77} & \roundedbis{1.77} & \roundedbis{1.82} & \multicolumn{1}{c|}{\textit{-0.22}} \\
\multicolumn{1}{|c}{} & \textbf{NER} & \cellcolor{red!25} \roundedbis{1.54} & \cellcolor{red!25} \roundedbis{1.62} & \cellcolor{red!25} \roundedbis{2.26} & \cellcolor{red!25} \roundedbis{2.27} & \roundedbis{1.76} & \roundedbis{1.72} & \roundedbis{1.73} & \cellcolor{red!25} \roundedbis{1.72} & \roundedbis{1.54} & \roundedbis{1.67} & \roundedbis{2.24} & \roundedbis{2.3} & \cellcolor{red!25} \roundedbis{1.72} & \roundedbis{1.77} & \cellcolor{green!25} \roundedbis{1.77} & \roundedbis{1.82} & \multicolumn{1}{c|}{\textit{-0.22}} \\ \hline

\multicolumn{1}{|c}{\multirow{2}{*}{\textbf{DEFT2020}}}
& \textbf{STS} & \roundedbis{1.41} & \roundedbis{1.45} & \roundedbis{2.27} & \roundedbis{2.24} & \roundedbis{1.42} & \roundedbis{1.45} & \roundedbis{1.43} & \roundedbis{1.45} & \cellcolor{green!25} \roundedbis{1.41} & \roundedbis{1.49} & \cellcolor{red!25} \roundedbis{2.24} & \cellcolor{red!25} \roundedbis{2.23} & \roundedbis{1.41} & \roundedbis{1.48} & \roundedbis{1.42} & \roundedbis{1.49} & \multicolumn{1}{c|}{\textit{-0.47}} \\
\multicolumn{1}{|c}{}  & \textbf{CLS} & \cellcolor{green!25} \roundedbis{1.21} & \roundedbis{1.26} & \cellcolor{red!25} \roundedbis{2.13} & \roundedbis{2.09} & \roundedbis{1.31} & \roundedbis{1.34} & \roundedbis{1.33} & \roundedbis{1.36} & \roundedbis{1.2} & \roundedbis{1.32} & \roundedbis{2.05} & \roundedbis{2.04} & \roundedbis{1.25} & \roundedbis{1.34} & \roundedbis{1.29} & \roundedbis{1.37} & \multicolumn{1}{c|}{\textit{-0.41}} \\ \hline

\multicolumn{1}{|c}{\multirow{1}{*}{\textbf{MorFITT}}}
& \textbf{CLS} & \roundedbis{1.38} & \roundedbis{1.44} & \roundedbis{2.45} & \cellcolor{red!25} \roundedbis{2.4} & \roundedbis{1.48} & \roundedbis{1.5} & \roundedbis{1.49} & \roundedbis{1.51} & \cellcolor{green!25} \roundedbis{1.37} & \roundedbis{1.5} & \roundedbis{2.35} & \roundedbis{2.33} & \roundedbis{1.46} & \roundedbis{1.55} & \roundedbis{1.48} & \roundedbis{1.57} & \multicolumn{1}{c|}{\textit{-0.82}} \\ \hline

\multicolumn{1}{|c}{\multirow{2}{*}{\textbf{E3C}}}
& \textbf{NER Clinical} & \cellcolor{green!25} \roundedbis{1.3} & \roundedbis{1.35} & \roundedbis{2.23} & \cellcolor{red!25} \roundedbis{2.17} & \roundedbis{1.48} & \roundedbis{1.48} & \roundedbis{1.5} & \roundedbis{1.49} & \roundedbis{1.29} & \roundedbis{1.43} & \roundedbis{2.22} & \roundedbis{2.18} & \roundedbis{1.48} & \roundedbis{1.55} & \roundedbis{1.49} & \roundedbis{1.56} & \multicolumn{1}{c|}{\textit{-0.59}} \\
 \multicolumn{1}{|c}{} & \textbf{NER Temporal} & \roundedbis{1.29} & \roundedbis{1.35} & \roundedbis{2.22} & \roundedbis{2.16} & \roundedbis{1.48} & \roundedbis{1.48} & \cellcolor{green!25} \roundedbis{1.48} & \roundedbis{1.49} & \roundedbis{1.29} & \roundedbis{1.43} & \roundedbis{2.22} & \cellcolor{red!25} \roundedbis{2.18} & \roundedbis{1.47} & \roundedbis{1.54} & \roundedbis{1.48} & \roundedbis{1.55} & \multicolumn{1}{c|}{\textit{-0.75}} \\ \hline

\multicolumn{1}{|c}{\multirow{1}{*}{\textbf{CLISTER}}}
& \textbf{STS} & \roundedbis{1.52} & \roundedbis{1.59} & \roundedbis{2.65} & \cellcolor{red!25} \roundedbis{2.57} & \roundedbis{1.73} & \roundedbis{1.72} & \roundedbis{1.74} & \cellcolor{green!25} \roundedbis{1.72} & \roundedbis{1.51} & \roundedbis{1.65} & \roundedbis{2.56} & \roundedbis{2.49} & \roundedbis{1.71} & \cellcolor{red!25} \roundedbis{1.77} & \roundedbis{1.71} & \roundedbis{1.77} & \multicolumn{1}{c|}{\roundedbis{-0.33}} \\ \hline

\multicolumn{1}{|c}{\multirow{2}{*}{\textbf{DEFT2021}}} 
 & \textbf{NER} & \roundedbis{1.31} & \roundedbis{1.37} & \roundedbis{2.26} & \cellcolor{red!25} \roundedbis{2.19} & \roundedbis{1.48} & \roundedbis{1.49} & \roundedbis{1.5} & \roundedbis{1.5} & \cellcolor{green!25} \roundedbis{1.31} & \roundedbis{1.44} & \roundedbis{2.19} & \roundedbis{2.15} & \roundedbis{1.48} & \roundedbis{1.55} & \roundedbis{1.49} & \roundedbis{1.56} & \multicolumn{1}{c|}{\textit{-0.88}} \\ 
\multicolumn{1}{|c}{} & \textbf{CLS} & \roundedbis{1.5} & \roundedbis{1.57} & \roundedbis{2.63} & \roundedbis{2.56} & \roundedbis{1.69} & \roundedbis{1.7} & \roundedbis{1.71} & \roundedbis{1.71} & \roundedbis{1.46} & \roundedbis{1.61} & \roundedbis{2.5} & \roundedbis{2.46} & \cellcolor{red!25} \roundedbis{1.64} & \roundedbis{1.72} & \roundedbis{1.66} & \cellcolor{green!25} \roundedbis{1.74} & \multicolumn{1}{c|}{\textit{-0.11}} \\ \hline

\multicolumn{1}{|c}{\multirow{4}{*}{\textbf{ESSAI}}}
 & \textbf{NER Spec} & \roundedbis{1.29} & \roundedbis{1.34} & \cellcolor{red!25} \roundedbis{2.2} & \roundedbis{2.14} & \roundedbis{1.42} & \roundedbis{1.43} & \roundedbis{1.45} & \roundedbis{1.45} & \roundedbis{1.29} & \roundedbis{1.41} & \roundedbis{2.21} & \roundedbis{2.16} & \cellcolor{green!25} \roundedbis{1.41} & \roundedbis{1.49} & \roundedbis{1.46} & \roundedbis{1.52} & \multicolumn{1}{c|}{\textit{-0.68}} \\
\multicolumn{1}{|c}{} & \textbf{POS} & \roundedbis{1.28} & \roundedbis{1.33} & \roundedbis{2.19} & \cellcolor{red!25} \roundedbis{2.13} & \roundedbis{1.41} & \roundedbis{1.42} & \roundedbis{1.44} & \roundedbis{1.44} & \roundedbis{1.28} & \roundedbis{1.41} & \roundedbis{2.19} & \cellcolor{red!25} \roundedbis{2.15} & \cellcolor{green!25} \roundedbis{1.4} & \roundedbis{1.48} & \roundedbis{1.44} & \roundedbis{1.51} & \multicolumn{1}{c|}{\textit{-0.61}} \\
\multicolumn{1}{|c}{}  & \textbf{NER Neg} & \roundedbis{1.28} & \roundedbis{1.33} & \roundedbis{2.19} & \cellcolor{red!25} \roundedbis{2.13} & \roundedbis{1.41} & \roundedbis{1.42} & \roundedbis{1.44} & \roundedbis{1.44} & \roundedbis{1.28} & \roundedbis{1.41} & \roundedbis{2.19} & \roundedbis{2.15} & \roundedbis{1.4} & \cellcolor{green!25} \roundedbis{1.48} & \roundedbis{1.44} & \roundedbis{1.51} & \multicolumn{1}{c|}{\textit{-0.69}} \\
\multicolumn{1}{|c}{} & \textbf{CLS} & \roundedbis{1.28} & \roundedbis{1.34} & \roundedbis{2.2} & \cellcolor{red!25} \roundedbis{2.14} & \roundedbis{1.42} & \roundedbis{1.43} & \roundedbis{1.45} & \roundedbis{1.46} & \roundedbis{1.28} & \roundedbis{1.41} & \cellcolor{green!25} \roundedbis{2.2} & \roundedbis{2.16} & \roundedbis{1.41} & \roundedbis{1.49} & \roundedbis{1.45} & \roundedbis{1.52} & \multicolumn{1}{c|}{\textit{-0.02}} \\ \hline

\multicolumn{1}{|c}{\multirow{2}{*}{\textbf{QUAERO}} }
 & \textbf{NER Medline} & \roundedbis{1.53} & \roundedbis{1.63} & \roundedbis{2.35} & \roundedbis{2.26} & \roundedbis{1.78} & \roundedbis{1.78} & \roundedbis{1.77} & \roundedbis{1.78} & \cellcolor{green!25} 
 \roundedbis{1.52} & \roundedbis{1.76} & \roundedbis{2.36} & \cellcolor{red!25} \roundedbis{2.35} & \roundedbis{1.77} & \roundedbis{1.89} & \roundedbis{1.76} & \roundedbis{1.89} & \multicolumn{1}{c|}{\textit{-0.77}} \\
\multicolumn{1}{|c}{} & \textbf{NER EMEA} & \roundedbis{1.3} & \roundedbis{1.34} & \roundedbis{2.14} & \roundedbis{2.12} & \roundedbis{1.44} & \roundedbis{1.46} & \roundedbis{1.49} & \roundedbis{1.51} & \roundedbis{1.3} & \cellcolor{green!25}   \roundedbis{1.39} & \roundedbis{2.06} & \roundedbis{2.04} & \roundedbis{1.45} & \roundedbis{1.51} & \roundedbis{1.5} & \cellcolor{red!25} \roundedbis{1.56} & \multicolumn{1}{c|}{\textit{-0.28}} \\ \hline

\multicolumn{1}{|c}{\multirow{3}{*}{\textbf{MANTRAGSC}} }
& \textbf{NER EMEA} & \roundedbis{1.33} & \roundedbis{1.40} & \roundedbis{2.47} & \roundedbis{2.41} & \cellcolor{green!25}   \roundedbis{1.49} & \roundedbis{1.51} & \roundedbis{1.5} & \roundedbis{1.52} & \roundedbis{1.32} & \roundedbis{1.43} & \cellcolor{red!25} \roundedbis{2.33} & \roundedbis{2.3} & \roundedbis{1.46} & \roundedbis{1.53} & \roundedbis{1.49} & \roundedbis{1.55} & \multicolumn{1}{c|}{\textit{-0.63}} \\
\multicolumn{1}{|c}{}  & \textbf{NER Medline} & \roundedbis{1.89} & \roundedbis{2.01} & \roundedbis{2.84} & \cellcolor{red!25} \roundedbis{2.7} & \roundedbis{2.06} & \roundedbis{2.13} & \roundedbis{2.14} & \roundedbis{2.14} & \roundedbis{1.89} & \roundedbis{2.09} & \roundedbis{2.84} & \roundedbis{2.78} & \roundedbis{2.06} & \roundedbis{2.22} & \cellcolor{green!25}   \roundedbis{2.1} & \roundedbis{2.22} & \multicolumn{1}{c|}{\textit{-0.64}} \\
 \multicolumn{1}{|c}{} & \textbf{NER Patents} & \roundedbis{1.54} & \roundedbis{1.59} & \cellcolor{green!25}   \roundedbis{2.34} & \roundedbis{2.3} & \roundedbis{1.61} & \roundedbis{1.63} & \roundedbis{1.59} & \roundedbis{1.62} & \roundedbis{1.43} & \roundedbis{1.52} & \roundedbis{2.20} & \roundedbis{2.20} & \roundedbis{1.50} & \roundedbis{1.58} & \roundedbis{1.51} & \cellcolor{red!25} \roundedbis{1.60} & \multicolumn{1}{c|}{\textit{0.06}} \\ \hline

 
 \hline
 \hline

\multicolumn{2}{|c|}{\multirow{1}{*}{\textbf{Average per model}} }
& \roundedbis{1,39} & \roundedbis{1,45} & \roundedbis{2,30} & \roundedbis{2,25} & \roundedbis{1,54} & \roundedbis{1,55} & \roundedbis{1,56} & \roundedbis{1,56} & \roundedbis{1,38} & \roundedbis{1,51} & \roundedbis{2,26} & \roundedbis{2,24} & \roundedbis{1,52} & \roundedbis{1,60} & \roundedbis{1,55} & \roundedbis{1,62} & \multicolumn{1}{c|}{\textit{-0.48}}  \\ \hline

\multicolumn{2}{|c|}{\multirow{1}{*}{\textbf{Relative Difference (\%)}} }
& \rounded{0.0} &  \rounded{4,53} & \rounded{65,87} & \rounded{61,78} & \rounded{11,21} & \rounded{11,78} & \rounded{12,34} & \rounded{12,59} & \rounded{-0,65} & \rounded{8,87} & \rounded{62,84} & \rounded{61,09} & \rounded{9,87} & \rounded{15,46} & \rounded{11,71} & \rounded{16,90} &  \\

\cline{1-18}

\end{tabular}
\caption{Average number of sub-word units per word for each tokenization strategy and data source training. Their Pearson correlation ($\rho$) with each task performance is reported (last column). Cells colored in red correspond to lower performing models, while those in green represent higher ones. The last row represents the relative difference in terms of average subwords per word compared to the NACHOS BPE without morpheme baseline. {\it w/o} and {\it w/} denote models without and with morphemes.}
\label{table:avg-subwords-words-pearson}
\end{table*}

\subsection{Impact of tokenization granularity}
\label{sec:impact-token-granularity}




To assess the impact of tokenization granularity, Table~\ref{table:avg-subwords-words-pearson} presents the average number of sub-word units per word for each tokenization strategy and data source used in the studied tasks. While deriving overarching conclusions from these results can be challenging, we calculated Pearson correlation ($\rho$) between models performances on the downstream tasks from Table~\ref{table:results} and the corresponding average number of sub-word units per word. These correlation scores range from $-1$ to $+1$, where $-1$ indicates a complete negative linear correlation, $0$ represents no correlation, and $+1$ signifies a strong positive correlation. In the context of tokenization, a negative correlation implies that fewer subword units are associated with higher scores, while a positive correlation suggests that more subword units are linked to higher scores.




In overall, we observe in Table~\ref{table:avg-subwords-words-pearson} an average $\rho$ correlation of $-0.48$ between tasks and models, indicating that, in general, higher performance scores tend to be associated with fewer subword units. To our knowledge, this is the first time such a correlation has been experimentally demonstrated. However, it's important to note that this correlation varies across the targeted tasks. Tasks like CLS show correlation close to zero, suggesting that they are less affected by the granularity of tokenization. In contrast, STS and sequence-to-sequence tasks, particularly NER, appear to be more influenced by tokenization granularity, likely due to their heavy reliance on immediate context for making predictions.

While the RoBERTa model's embeddings capture semantic meaning and the encoder module captures contextual information~\cite{rogers-etal-2020-primer}, we aimed to determine whether the observed correlations are attributed to a specific part of this architecture. To investigate this, we isolated and froze the embeddings and/or encoder of our BERT-based model, based on the NACHOS SentencePiece, during fine-tuning for various tasks. The experimental approach, as detailed in Table~\ref{table:frozen}, involved several stages. Initially, we established a baseline for each task with no frozen components. Subsequently, we conducted experiments by freezing only the embedding layer, only the encoder, and both the embeddings and encoders. Our findings indicate a stronger dependence on RoBERTa's encoder for tasks such as POS tagging and STS, in contrast to other tasks, which corroborate the context dependency as an explanation to the correlation scores between segmentation granularity and models performances for these tasks but not for NER.

\begin{table}[!htb]
\setlength\extrarowheight{6.0pt}
\resizebox{\columnwidth}{!}{%
\begin{tabular}{|l|cccc|}
\hline
\textbf{} &
\textbf{\begin{tabular}[c]{@{}c@{}}CAS\\ POS\end{tabular}} &
\textbf{\begin{tabular}[c]{@{}c@{}}PxCorpus\\ NER\end{tabular}} &
\textbf{\begin{tabular}[c]{@{}c@{}}PxCorpus\\ CLS\end{tabular}} &
\textbf{\begin{tabular}[c]{@{}c@{}}CLISTER\\ STS\end{tabular}} \\

\hline

\textbf{ \includegraphics[width=1em]{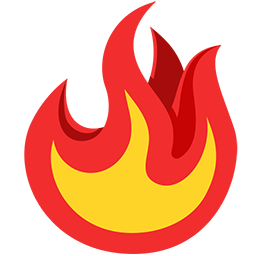} Full Fine-tuning}           & 97.10 & 96.10 & 94.82 & 0.61 \\

\hdashline

\textbf{ \includegraphics[width=1em]{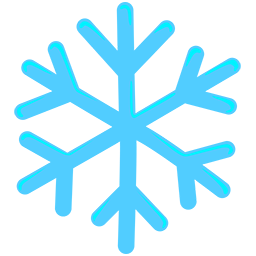} Embedding}           & 97.03 $\downarrow$ \scalebox{0.66}{0.07} & 96.10 $\uparrow$ \scalebox{0.66}{0.00} & 94.73 $\downarrow$ \scalebox{0.66}{0.09} & 0.62 $\uparrow$ \scalebox{0.66}{1.63} \\

\textbf{ \includegraphics[width=1em]{images/snowflake_2744.png} Encoder}             & 65.97 $\downarrow$ \scalebox{0.66}{32.05}  & 83.95 $\downarrow$ \scalebox{0.66}{12.64} & 84.78 $\downarrow$ \scalebox{0.66}{10.58} & 0.45 $\downarrow$ \scalebox{0.66}{26.22} \\

\textbf{ \includegraphics[width=1em]{images/snowflake_2744.png} Embedding + Encoder} & 60.04 $\downarrow$ \scalebox{0.66}{38.16} & 79.62 $\downarrow$ \scalebox{0.66}{17.14} & 84.78 $\downarrow$ \scalebox{0.66}{10.58} & 0.44 $\downarrow$ \scalebox{0.66}{27.86} \\ \hline

\end{tabular}%
}
\caption{Performance and relative loss (in \%) of the PLMs based on SentencePiece NACHOS without morpheme with parts of the models being frozen.}
\label{table:frozen}
\end{table}

As shown in Table~\ref{table:avg-subwords-words-pearson}, higher performance scores are associated with fewer subword units. To gain a linguistic perspective on how tokenization strategies behave, we analyzed the segmentation of 150 biomedical terms equally distributed across cardiology, dermatology, obstetric-gynecology, and ophthalmology, as presented in Table~\ref{table:lexique-150}. Most models, except for those using SentencePiece NACHOS, struggle to precisely align with the official morphological segmentation established by the Académie Française (French Academy). However, upon closer examination, it is evident that these models often come very close to the desired segmentation. While the segmentations may exhibit slight variations, such as the relocation of a letter from one token to another, they maintain the same number of tokens as the official morphological segmentation. This observation is further supported when we analyze actual tokenizer outputs (see Table~\ref{table:tokenization-examples}) and assess the segmentation statistics in Table~\ref{table:lexique-150}. For example, BPE NACHOS tokenizes the term "ophtalmoscope" into the units "ophtalm oscope," whereas the morphological segmentation should be "ophtalmo scope," a segmentation achieved by its morpheme-enriched counterpart.

\begin{table}[!htb]
\scriptsize
\centering
\setlength\tabcolsep{2pt}
\setlength\extrarowheight{1.5pt}
\begin{tabular}{cllcc|cc|}

\cline{6-7}
 &  &  &  &  & \multicolumn{2}{c|}{\textbf{Type of errors}}  \\ 
\cline{4-7}
 &  &\multicolumn{1}{c|}{}  & \textbf{EM}* & \textbf{Exact \# Tok.} & \textbf{Under Seg.} & \textbf{Over Seg.} \\ \hline

\multicolumn{1}{|l}{\multirow{8}{*}{\textbf{BPE}}} & \multirow{2}{*}{\textbf{\textit{NACHOS}}} &  \multicolumn{1}{c|}{\textit{w/o}}  &   21.3  &  41.3  &  9.3  &  49.3 \\
\multicolumn{1}{|c}{} &  & \multicolumn{1}{c|}{\textit{w/}}  &  34.6  &  50.0  &  6.0  &  44.0 \\
\multicolumn{1}{|c}{} & \multirow{2}{*}{\textbf{\textit{PubMed}}} &  \multicolumn{1}{c|}{\textit{w/o}} &  2.6  &  12.0  &  2.6  &  85.3 \\
\multicolumn{1}{|c}{} &  & \multicolumn{1}{c|}{\textit{w/}}  &  17.3  &  28.6  &  2.6  &  68.6 \\
\multicolumn{1}{|c}{} & \multirow{2}{*}{\textbf{\textit{CC100}}} &  \multicolumn{1}{c|}{\textit{w/o}} &  8.0  &  28.0  &  2.6  &  69.3 \\
\multicolumn{1}{|c}{} &  & \multicolumn{1}{c|}{\textit{w/}}  &  23.3  &  38.6  &  2.6  &  58.6 \\
\multicolumn{1}{|c}{} & \multirow{2}{*}{\textbf{\textit{Wikipedia}}} &  \multicolumn{1}{c|}{\textit{w/o}} &  8.6  &  24.6  &  3.3  &  72.0 \\
\multicolumn{1}{|c}{} &  & \multicolumn{1}{c|}{\textit{w/}}  &  22.0  &  36.6  &  4.6  &  58.6 \\

\hline

\multicolumn{1}{|l}{\multirow{8}{*}{\textbf{SP}}} & \multirow{2}{*}{\textbf{\textit{NACHOS}}} & \multicolumn{1}{c|}{\textit{w/o}} &  56.6  &  74.6  &  7.3  &  18.0 \\
\multicolumn{1}{|c}{} &  & \multicolumn{1}{c|}{\textit{w/}}  &  61.3  &  70.6  &  2.6  &  26.6 \\
\multicolumn{1}{|c}{} & \multirow{2}{*}{\textbf{\textit{PubMed}}} &  \multicolumn{1}{c|}{\textit{w/o}} &  14.6  &  26.6  &  2.6  &  70.6 \\
\multicolumn{1}{|c}{} &  & \multicolumn{1}{c|}{\textit{w/}}  &  32.0  &  42.0  &  2.6  &  55.3 \\
\multicolumn{1}{|c}{} & \multirow{2}{*}{\textbf{\textit{CC100}}} &  \multicolumn{1}{c|}{\textit{w/o}} &  24.0  &  42.0  &  4.0  &  54.0 \\
\multicolumn{1}{|c}{} &  & \multicolumn{1}{c|}{\textit{w/}}  &  36.6  &  49.3  &  2.6  &  48.0 \\
\multicolumn{1}{|c}{} & \multirow{2}{*}{\textbf{\textit{Wikipedia}}} &  \multicolumn{1}{c|}{\textit{w/o}} &  18.0  &  42.0  &  3.3  &  54.6 \\
\multicolumn{1}{|c}{} &  & \multicolumn{1}{c|}{\textit{w/}}  &  34.0  &  54.0  &  4.6  &  41.3 \\

\hline

\end{tabular}%
\caption{The average Exact Match (EM*) and portion of terms aligned with the official segmentation length (Exact \# Tok.), both in \%, are based on the gold segmentation from 150 biomedical terms. Both last columns are referring to the portion of terms suffering from under and over segmentation. {\it w/o} and {\it w/} denote without and with morphemes respectively. SP stands for SentencePiece.}
\label{table:lexique-150}
\end{table}

In Table~\ref{table:lexique-150}, we observed various types of errors in segmentation, with the most common issue being over-segmentation of units that are not present in our biomedical lexical morphemes list. This over-segmentation results in smaller, more numerous, and sparser tokens, which can impact the efficiency of pre-training. The reduced frequency of tokens and the faster filling of RoBERTa's 512-token context window with less meaningful tokens can be problematic.

\begin{table}[!t]
\resizebox{\columnwidth}{!}{%
\setlength\extrarowheight{3.0pt}
\centering
\begin{tabular}{|l|lll|}
\hline
\textbf{Base}       & \textit{cancérigène}    & \textit{ophtalmoscope}         & \textit{angiographie}     \\ \hline
\textbf{Correct}      & \textbf{cancér i gène}  & \textbf{ophtalmo scope}       & \textbf{angio graphie}    \\ \hline

\textbf{BPE Wiki}   & c anc éri gène & oph tal mos cope       & ang i ographie  \\ 
\textbf{BPE PubMed}   & can c é rig è ne  & o ph tal m oscope        & angi ograph ie  \\ 
\textbf{BPE NACHOS} & cancé rig ène  & ophtalm oscope & angiographie  \\ 
\textbf{SentencePiece NACHOS}             & cancérigène         & ophtalm oscope & angiographie     \\ 
\textbf{BPE NACHOS \texttt{+Morpheme} }            & \textbf{cancér i gène} & \textbf{ophtalmo scope}        & \textbf{angio graphie} \\ 
\textbf{SentencePiece NACHOS \texttt{+Morpheme} } & \textbf{cancér i gène} & \textbf{ophtalmo scope}        & \textbf{angio graphie}      \\ \hline
\end{tabular}%
}
\caption{Instances of tokenization juxtaposed with their correct segmentation.}
\label{table:tokenization-examples}
\end{table}

Finally, Table~\ref{table:lexique-150} reveals an interesting distinction between BPE and SentencePiece using NACHOS training data. SentencePiece outperforms BPE in achieving segmentations that closely resemble correct ones, both in terms of the number of tokens and their semantic accuracy. SentencePiece excels at matching correct segmentations, particularly for medical terminology, in 56.6\% of cases without morphemes and 61.3\% when morphemes are used, while BPE NACHOS achieves only 34.6\% accuracy.

\subsection{Impact of morphemes}
\label{sec:morpheme-impact}

One of our primary objectives was to approximate the correct morphological segmentation of words in the French biomedical language. Our analysis reveals that tokenizers, such as BPE and SentencePiece trained on NACHOS, enriched with morphemes, can often achieve this goal. Notably, SentencePiece NACHOS enriched with morphemes achieved the best performance, with a 61.3\% exact match. Our morpheme-enriched approach offers the advantage of obtaining a tokenization that closely resembles what could be achieved through a complex rule-based method. This approach is easily adaptable to other languages with a list of lexical morphemes and similar principles.

As shown in Table~\ref{table:results}, the introduction of morphemes ({\it w/}) may lead to performance enhancements in approximately 25\% of the studied downstream tasks. However, it is noteworthy that the best results are primarily achieved by classical statistical tokenizers, BPE and SentencePiece, when not using morphemes, and when trained on our biomedical-specific data, NACHOS. This observation is intriguing because NACHOS-based tokenizers inherently contain a higher proportion of morphemes, as shown in Table~\ref{table:recouvrement}, which presents the portion of correct morphemes already present in the tokenizers without introducing additional morphological information based on their length ranges. This suggests that introducing morphemes and other forms of morphological knowledge, such as grammatical endings, may have a more substantial impact in contexts that do not align directly with the target domains and languages. However, we can note that the results of this method are inconsistent and do not ensure an overall performance boost across all models or tasks.

\begin{table}[!htb]
\scriptsize
\centering
\setlength\tabcolsep{6.5pt}
\setlength\extrarowheight{1.5pt}
\begin{tabular}{cc|cccc|}
\cline{3-6}
&                    & \multicolumn{4}{c|}{\textbf{Coverage of the morphemes (\%)}} \\ \hline
\multicolumn{1}{|c}{\textbf{Tokenizer}} & \textbf{Source} & \textbf{1 - 3} & \textbf{4 - 6} & \multicolumn{1}{c|}{\textbf{7 - 10}} & \textbf{Global} \\ \hline
\multicolumn{1}{|c}{\multirow{4}{*}{\textit{\textbf{BPE}}}} & \textit{\textbf{NACHOS}}    & 83.33   & 45.38   & \multicolumn{1}{c|}{31.00}   & 47.23   \\
\multicolumn{1}{|c}{}                              & \textit{\textbf{PubMed}}    & 65.15   & 39.32   & \multicolumn{1}{c|}{15.00}   & 38.06   \\
\multicolumn{1}{|c}{}                              & \textit{\textbf{CC100}}     & 78.78   & 34.46   & \multicolumn{1}{c|}{7.00}    & 34.77   \\
\multicolumn{1}{|c}{}                              & \textit{\textbf{Wikipedia}} & 87.87   & 34.95   & \multicolumn{1}{c|}{10.00}   & 36.67   \\ \hline
\multicolumn{1}{|c}{\multirow{4}{*}{\textit{\textbf{SP}}}} & \textit{\textbf{NACHOS}}    & 83.33   & 41.01   & \multicolumn{1}{c|}{28.00}   & 43.59   \\
\multicolumn{1}{|c}{}                              & \textit{\textbf{PubMed}}    & 60.60   & 37.13   & \multicolumn{1}{c|}{14.00}   & 35.81   \\
\multicolumn{1}{|c}{}                              & \textit{\textbf{CC100}}     & 83.33   & 34.70   & \multicolumn{1}{c|}{8.00}    & 35.64   \\
\multicolumn{1}{|c}{}                              & \textit{\textbf{Wikipedia}} & 93.93   & 37.37   & \multicolumn{1}{c|}{12.0}    & 39.44   \\ \hline
\end{tabular}%
\caption{Percentage of the morphemes already present in the tokenizers vocabularies per range of morphemes lengths. SP stands for SentencePiece.\label{table:recouvrement}}
\end{table}

Furthermore, it is worth noting that morphemes are often already present in the tokenizers in their complete form, as illustrated in Table~\ref{table:recouvrement}, or with minor modifications based on token probabilities, as shown in Table~\ref{table:tokenization-examples}. Notably, tokenizers based on NACHOS contain a significantly higher percentage of morphemes, with 47.23\% for BPE and 43.59\% for SentencePiece. Conversely, the source with the fewest morphemes is CC100, with percentages of 34.77\% for BPE and 35.64\% for SentencePiece. This observation aligns with the fact that CC100 has fewer connections to both the target language and domain.

In general, we observe that despite the significant improvement in segmentation quality (as shown in Table~\ref{table:lexique-150}), tokenizers enriched with morphemes do not exhibit a strong correlation with the results achieved in downstream tasks, as evident in Table~\ref{table:results}. The ability to deliver satisfactory results despite encountering suboptimal segmentations, as seen in the case of PubMed, which frequently over-segments words, underscores the robustness of RoBERTa's architecture in handling noise and its capacity to compensate for such challenges.




\subsection{Impact of data sources}
\label{sec:impact-data-sources}

As indicated in Table~\ref{table:results}, the average performance across tasks demonstrates a significant impact of the training data source on the results obtained by the models. It becomes apparent that using data that is more suitable for the target language, even if it originates from various domains such as Wikipedia and CC100, is more effective than utilizing data from the target domains but from a different language. This is particularly evident in the CLS, NER, and STS tasks, where BPE PubMed achieves an average of 70.16\% for classification, 0.63 MSE for STS, and 62.62\% for NER, whereas CC100 outperforms with 74.14\%, 0.67 MSE, and 64.62\%, respectively.

The decrease in performance from PubMed can be attributed to over-segmentation, as seen in Table~\ref{table:avg-subwords-words-pearson}. This over-segmentation is primarily due to the significant differences between the data used to build the tokenizer and the language of the model's pre-training. These differences stem from distinct lexicons, writing styles, and morphological structures in French compared to English, particularly for specialized words like "Péricardite" (French) and "Pericarditis" (English), or "Orthophoniste" (French) and "Speech Therapist" (English). Furthermore, variations in alphabets, such as special French characters like "é" or "è," can lead to token sparsity when encountered in positions not seen during tokenizer construction on PubMed. This results in a lack of both language and domain-specific information for French, as only limited tokens can be used to form sentences.





Some data sources are surprisingly less affected by the introduction of morphemes. For instance, the CC100 source is not positively impacted by morphemes, despite having a lower proportion of morphemes in its original version, as shown in Table~\ref{table:recouvrement}. This behavior may be explained by the increased granularity introduced by morphemes, which reduces the probabilities of other tokens appearing. This can lead to a poorer representation of words.

\section{Conclusion}



In this study, we conducted a comprehensive investigation into the influence of various word tokenization strategies within a BERT-based masked language model across diverse French biomedical NLP tasks. Notably, we observed that existing methods for tokenizing biomedical text often fall short of aligning with morphological rules and how humans learn these specialized terms. This suboptimal segmentation can impact the agglutinating nature of biomedical terminology. To assess the effects of this segmentation on downstream applications, we developed a set of novel biomedical tokenizers that adhere more closely to morphological rules. These tokenizers combine various automatic tokenization approaches and vocabularies to enrich segmentation with morphemes. We employed these enhanced tokenizers in the pre-training of multiple RoBERTa-based models, which we then evaluated across a wide array of 23 French biomedical tasks, including POS, NER, STS, and CLS.

Our findings show that integrating morphemes into automatic tokenization approaches can achieve parity or improve performance in certain tasks, such as NER and POS tagging. However, this enhancement is not consistent across all tasks. While there is a correlation between segmentation granularity and downstream task performance, we also observe that pre-training processes exhibit robustness to suboptimal tokenization, yielding surprisingly good results even with very short and sparse subword units. To conclude, our study reveals that achieving optimal tokenization involves a combination of factors, including minimizing word segmentation and having access to domain-specific data in the target language.



\section{Acknowledgements}

This work was performed using HPC resources from GENCI-IDRIS (Grant 2022-AD011013061R1 and 2022-AD011013715). This work was financially supported by ANR MALADES (ANR-23-IAS1-0005) and Zenidoc.




\section{Bibliographical References}\label{sec:reference}
\bibliographystyle{lrec-coling2024-natbib}
\bibliography{lrec-coling2024-example}
\section{Language Resource References}\label{lr:ref}
\bibliographystylelanguageresource{lrec-coling2024-natbib}
\bibliographylanguageresource{languageresource}

\end{document}